\pdfoutput=1

\documentclass[11pt]{article}

\usepackage[]{acl}

\usepackage{times}
\usepackage{latexsym}

\usepackage[T1]{fontenc}

\usepackage[utf8]{inputenc}

\usepackage{microtype}

\usepackage{inconsolata}
\usepackage{booktabs}
\usepackage{spverbatim}
\usepackage{listings}
\usepackage{amsmath}
\usepackage{comment}
\usepackage{enumitem}
\usepackage{array}

\usepackage{subcaption}
\usepackage{graphicx}

\title{Towards Faithful and Plausible Natural Language Explanations for Image Classification: A Pipeline Approach}
\title{Faithful and Plausible Natural Language Explanations for Image Classification: A Pipeline Approach}

\author{Adam Wojciechowski$^{1,3}$ \and Mateusz Lango$^{1,2}$ \and Ondřej Dušek$^2$ \\
  $^1$Poznan University of Technology,  Faculty of Computing and Telecommunications, Poland  \\
  $^2$Charles University, Faculty of Mathematics and Physics, Prague, Czechia\\
  $^3$Samsung AI Center Warsaw, Poland \\
   \texttt{a.wojciecho4@samsung.com, \{lango,odusek\}@ufal.mff.cuni.cz} \\}

\usepackage{graphicx}

\usepackage{tikz-dependency}

\usepackage[normalem]{ulem}

\def\ODdel#1{\bgroup\markoverwith{\textcolor{purple!60}{\rule[0.4ex]{2pt}{3pt}}}\ULon{#1}}

\def\MLdel#1{\bgroup\markoverwith{\textcolor{blue!60}{\rule[0.4ex]{2pt}{3pt}}}\ULon{#1}}

\begin{document}
\maketitle


\begin{abstract}
Existing explanation methods for image classification struggle to provide faithful and plausible explanations. 
This paper addresses this issue by proposing a post-hoc natural language explanation method that can be applied to any CNN-based classifier without altering its training process or affecting predictive performance. 
By analysing influential neurons and the corresponding activation maps, the method generates a faithful description of the classifier's decision process in the form of a structured meaning representation, which is then converted into text by a language model. 
Through this pipeline approach, the generated explanations are grounded in the neural network architecture, providing accurate insight into the classification process while remaining accessible to non-experts. 

Experimental results show that the NLEs constructed by our method are significantly more plausible and faithful than baselines. In particular, user interventions in the neural network structure (masking of neurons) are three times more effective.
\end{abstract}

\section{Introduction}

Despite remarkable advances in computer vision, the deployment of image classification systems, especially in critical domains, poses significant challenges.
One of them is the opacity of deep models and the difficulty of providing reliable explanations for their predictions~\cite{doshi2017accountability}. 

Therefore, several types of explanation methods have been proposed, including various forms of saliency maps~\cite{selvaraju2017grad}, feature importances~\cite{lime}, concept-based explanations~\cite{chen2019looks}, counterfactual explanations~\cite{vermeire2022explainable}, etc.
A particularly interesting form of explaining predictions is offered by natural language explanation (NLE) techniques~\citep{NIPS2018_8163,wu-mooney-2019-faithful}. 
Such explanations are not only 
understandable by non-expert users, but can also be used to support conversations with the user in a dialogue system~\cite{raczynski2023problem}.


\begin{figure*}
    \includegraphics[width=\textwidth]{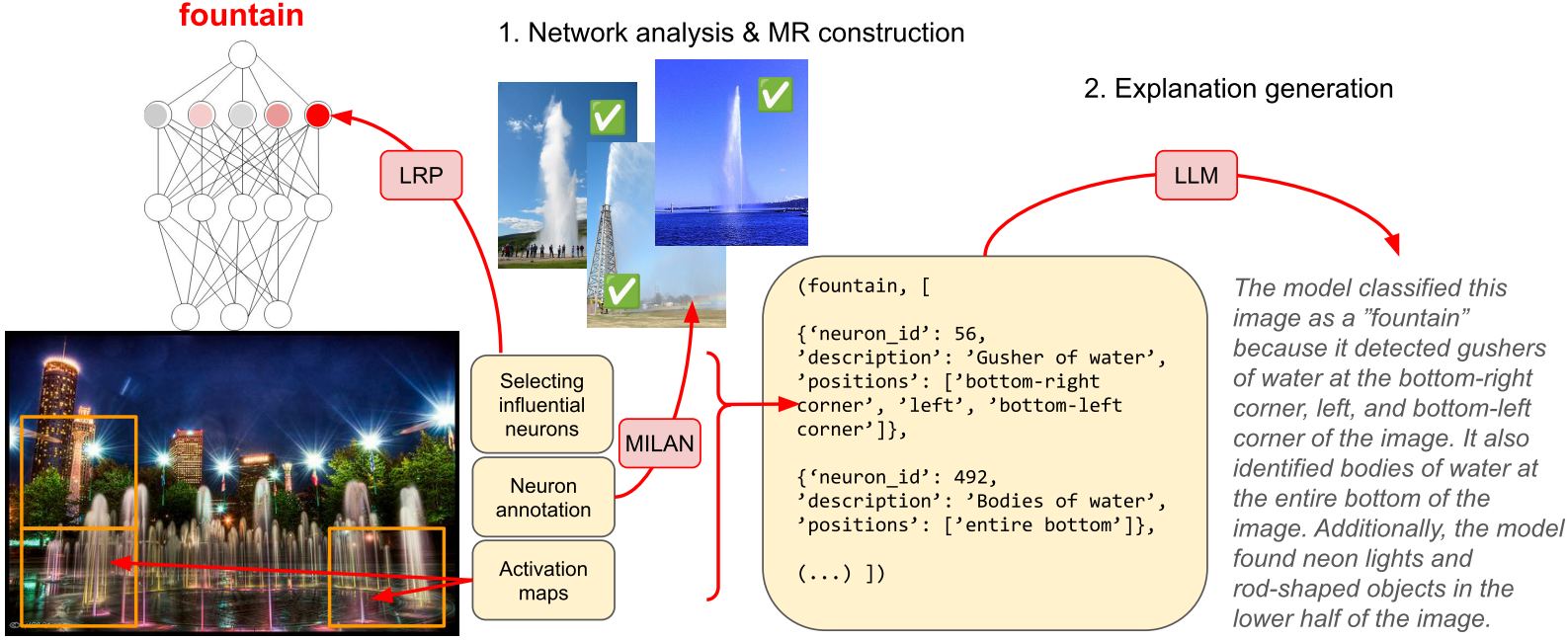}
    \caption{Overview of the presented approach. Note that the information provided in the text is supported by the model’s internal decision process, e.g. there is a convolutional filter specialized in the detection of "gushes of water", which was strongly activated at the mentioned positions of the image.}
    \label{fig:ex1}
\end{figure*}

There are two critical properties of an explanation: \emph{faithfulness} and \emph{plausibility}~\cite{jacovi-goldberg-2020-towards,atanasova-etal-2023-faithfulness}. 
A faithful explanation should accurately reflect the inner workings of the
system and provide information on the real reasons why the model reached a certain decision. 
Plausibility then refers to how convincing the explanation appears to the user.

In the case of NLE, obtaining high plausibility is straightforward, as textual explanations are usually human-friendly~\cite{gurrapu2023rationalization}, but achieving faithfulness is challenging. 
In the context of image classification, image captioning methods offer plausible but unfaithful NLEs~\cite{xu2015show,ai4030033}.
Some methods try to improve faithfulness by conditioning generation on both the predicted class and image features~\cite{hendricks2016generating,10.1007/978-3-030-01216-8_35,marasovic-etal-2020-natural, sammani2022nlx}, but the faithfulness provided is still limited as the model is not aware of the classifier's decision process.
Other methods train image classifiers to jointly predict the class and visual rationales, and generate explanations based on them \cite{9534269,xray}. However, the rationales are predicted independently of the class and do not participate in the classifier's decision process.
Most importantly, such methods change the training procedure and the architecture of the classifier, often affecting classification performance.

In this paper, we propose a post-hoc natural language explanation method for image classification that can be used with any standard convolutional neural network  (CNN) classifier. 
To illustrate the classification decision process, the method analyses which neurons of the CNN were most influential in reaching a given decision, and which regions of the image caused them to activate.
For each influential neuron, a neuron annotation method computes visual pattern exemplars and produces a short (one phrase) description of a pattern that the neuron detects.
The information gathered in this process serves to create a simple meaning representation, which is converted to natural-language text by a large language model (LLM).

Our simple pipeline method (see Fig.~\ref{fig:ex1} and Sect.~\ref{sec:method}) produces explanations that are directly grounded in the classifier's network architecture, but without interfering with its training process or affecting its predictive performance. 
It also does not need any gold-standard training explanations.
The provided NLEs reflect the process underlying classification by specifying the most influential neurons computed by well-established explainable AI methods.
At the same time, the final text is fluent and easy to understand.
This results in NLEs that are both plausible and highly faithful -- significantly more so than baselines, as demonstrated in Sect.~\ref{sec:experiments}.
Our experimental code is publicly available.\footnote{\url{https://github.com/wojciechowskiofficial/FLEX}}

\section{Method}
\label{sec:method}

Our NLE generation approach uses two processing steps, described below: meaning representation (MR) construction and MR-to-text conversion.

\subsection{Meaning representation construction}
We first produce a meaning representation  
in the form of a JSON object, containing information about the neurons responsible for a given classifier prediction (why?), what patterns those neurons detected (what?), and in which parts of the image they were activated (where?).
The MR includes the predicted class and the list of most influential neurons, each represented by
(1) description -- a phrase describing the pattern that most excites the neuron (convolutional filter), (2) positions -- list of coarse-grained image positions (e.g., “bottom-right corner”) where the neuron was activated. 
An example of a MR is provided in Fig.~\ref{fig:ex1}.


MR construction starts by storing all neuron activations from the given CNN-based classifier for the prediction to explain.
Next, the most important neurons are selected, annotated with a description, and tied to an image region, as follows:

\paragraph{Selecting the most influential neurons}
To pick the most relevant neurons, we apply the well-established Layer-wise Relevance Propagation (LRP) method \cite{Bach2015OnPE}. 
LRP performs a backward pass through the classifier network to establish the influence of each neuron to the final prediction (see App.~\ref{app:lrp} for formulas).
We select $k$ neurons with the highest LRP 
scores (with $k$ being a parameter controlling the brevity-detail tradeoff). 
 
\paragraph{Neuron annotation}
We adopt the MILAN neuron annotator~\cite{milan} to generate descriptions of selected neurons.
MILAN first finds images in the classifier training set 
that make a given neuron highly activated 
\cite{bau2017network}.
These exemplar images 
are used to generate a description of the pattern that this neuron detects.

Note that although the last step of MILAN is essentially image captioning, it does not affect the faithfulness of NLEs produced by the pipeline, as long as its output is of sufficient quality. 
The images that illustrate a neuron's decision process are computed by analysing its activations, and the captioning is only used to convert the result into text. 

\paragraph{Establishing image regions for neuron activation}
The neuron's raw activation map
is divided into a 3$\times$3 grid 
with manually assigned labels such as `top-left corner', `top', `top-right corner', etc. 
We then select
all grid cells where the neuron's activation exceeded half of its maximum value.
We apply several substitution rules (see App.~\ref{app:rules}) to make the list of cells shorter and more human-readable. 

\subsection{Explanation generation}
The second step of our method is converting the faithful MR created above
into a user-friendly text.
As we do not have any gold-standard explanation texts
, the task is performed by prompting a large language model (LLM). 

We instructed the model to (1) produce fluent text, (2) summarise the content of the MR (e.g.~if two neurons detect similar patterns, they can be combined in the text), (3) prioritise readability, (4) come up with its own formulation of spatial positions to improve fluency. We  also provide one handcrafted MR-to-text conversion example. 
The 
prompt is shown in App.~\ref{app:prompt}.
LLMs could in theory hallucinate and thus reduce the explanations' faithfulness.
However, in Section~\ref{sec:experiments} we show experimentally that current LLMs are reliable enough to produce useful explanations. 

\section{Experimental evaluation}
\label{sec:experiments}

\subsection{Experimental setup}

\paragraph{Dataset} All experiments were performed on the ImageNet dataset.
The classifier was trained on the training set and our explanation method was run to explain predictions made on the validation data.\footnote{Annotated ImageNet test set is not publicly available.}

\paragraph{Models} 
We experiment with explaining the predictions of the smallest CNN classifier from the popular ResNet family: ResNet18~\cite{DBLP:journals/corr/HeZRS15}.\footnote{
It reaches only a 41.4\% accuracy on the validation set, but high classification performance is not the goal of our study.} 
We fill our MRs with $k=10$ top neurons indicated by LRP from the Captum library~\cite{kokhlikyan2020captum} and annotate them using MILAN's original implementation. %
As the LLM for the MR-to-text conversion, we employ GPT-4~\citep[\texttt{gpt-4-0613};][]{openai2023gpt4}.

\paragraph{Baselines}
We compare to the following methods: 
\begin{itemize}[itemsep=0mm,parsep=1mm,topsep=1mm,leftmargin=3mm,labelwidth=3mm]
    \item
    Show, attend and tell (SAT) by~\citet{xu2015show} is an  image captioning method 
    used for explaining predictions~\cite{ai4030033}. 
\item 
    NLX-GPT~\cite{sammani2022nlx} is an explainable visual question-answering method that produces NLEs with an encoder-decoder architecture that combines CNN with a transformer-based language model.
\end{itemize}

\begin{table*}[t]
\small\centering
\begin{tabular}{l|ccc|ccc|ccc}
\toprule
                                 & \multicolumn{3}{c|}{Experts}             & \multicolumn{3}{c|}{Non-experts}        & \multicolumn{3}{c}{Overall}             \\
                                 & SAT           & NLX-GPT & Ours          & SAT          & NLX-GPT & Ours          & SAT           & NLX-GPT & Ours          \\\midrule
Fluency                          & \textbf{4.70}  & 4.12    & 4.64          & 3.64         & 2.80     & \textbf{3.70}  & \textbf{4.17} & 3.46    & \textbf{4.17} \\
Comprehensibility                & \textbf{4.94} & 4.42    & 4.18          & \textbf{3.70} & 2.88    & 3.24          & \textbf{4.32} & 3.65    & 3.71          \\
Plausibility (convincing)        & 2.16          & 2.28    & \textbf{3.44} & 2.00          & 2.22    & \textbf{2.70}  & 2.08          & 2.25    & \textbf{3.07} \\
Plausibility (explanatory) & 1.74          & 2.14    & \textbf{3.40}  & 2.14         & 2.28    & \textbf{2.94} & 1.94          & 2.21    & \textbf{3.17} \\
Overall quality                  & 2.12          & 2.40     & \textbf{3.46} & 1.94         & 2.02    & \textbf{2.54} & 2.03          & 2.21    & \textbf{3.00} \\\bottomrule
\end{tabular}
\caption{The results of a human evaluation experiment in which NLEs provided by different methods were evaluated on 5 factors.
 The overall inner-annotation agreement is 0.53 as measured by Krippendorff's alpha.}
\label{tab:res1}
\end{table*}

\subsection{Are the output explanations plausible?}

To assess the plausibility 
of generated explanations,
we conducted a small-scale manual annotation experiment.
We recruited ten annotators: five non-experts hired on the Prolific platform and five experts with at least one published paper on explainable AI. 
Each annotator was presented with 30 image-explanation pairs (300 in total) and asked to rate on a scale of 1-5 
whether the explanations were (1) fluent, (2) easy to understand (comprehensible), (3) convincing, and (4) insightful for the underlying decision process.\footnote{Note that the question on understanding the decision process does not measure faithfulness, but the user's subjective opinion on whether they understand how the model works.} The overall quality of the explanations was also rated (see App.~\ref{app:plaus}).

The results are presented in Tab.~\ref{tab:res1} and examples of generated NLEs can be found in Tab.~\ref{tab:case_study_table} (see App.~\ref{app:examples} for more).
Our method obtains the highest overall quality according to both experts and non-experts.
It also produces the most plausible explanations (most convincing and insightful). 
Since the baselines produce much shorter explanations, 
it is not surprising that our longer explanations are a bit more difficult to understand. 
Interestingly, experts generally give higher ratings than non-experts
for all methods and all factors except for providing insight into the decision process. 
Here, experts rate the baselines lower than non-experts, but they consistently rate the explanations provided by our pipeline higher.
The improvements of our method over baselines are statistically significant on both plausibility measures and overall quality. For fluency, 
our method is indistinguishable from SAT (see details in App.~\ref{app:stat}).

\subsection{Are the output explanations faithful?}

The faithfulness of the generated explanations is assessed through two intervention experiments: (1) checking if rationales from NLEs change the prediction by masking parts of input images, (2) influencing the network prediction by masking influential neurons.
We further assess the stability and diversity of the explanations
for our method, and we directly evaluate the reliability of our MR-to-text conversion.



\begin{table}[]
\small
\begin{tabular}{lllllll}
\toprule
        & \multicolumn{2}{l}{Covering} & \multicolumn{2}{l}{Highlighting} & \multicolumn{2}{l}{Neuron mask.\hspace{-1mm}} \\
        & c.f.$\uparrow$        & $\Delta$p$\uparrow$        & c.f.$\downarrow$          & $\Delta$p$\downarrow$        & c.f.$\uparrow$     & $\Delta$p$\uparrow$    \\\midrule
SAT     & 0.50           & 0.26        & 0.80            & 0.38         & 0.20         & 0.06      \\
NLX-GPT\hspace{-2mm} & 0.60           & 0.30        & 0.84            & 0.40         & 0.19         & 0.07      \\
Ours    & \textbf{0.88}           & \textbf{0.46}        & \textbf{0.66}            & \textbf{0.26}         & \textbf{0.66}         & \textbf{0.34}   \\\bottomrule
\end{tabular}
\caption{The results of three intervention experiments: percentage of examples for which user intervention resulted in a class flip (c.f.) and the average drop of probability of the predicted class ($\Delta$p).}
\label{tab:intervention}
\end{table}

\begin{table}[]
\small
\setlength{\tabcolsep}{4.5pt}
\begin{tabular}{lrrrr}
\toprule
                           & \multicolumn{1}{c}{BLEU} & \multicolumn{1}{c}{MET.} &c.f.&$\Delta$p.\\
                           \midrule
\hspace{-1mm}Intra-set stability (5\% noise)\hspace{-2mm} & 41.33           & 0.610         & 0.32  &  0.153  \\
\hspace{-1mm}Intra-set stability\,(20\% noise)\hspace{-5mm}  & 30.87         & 0.521        & 0.81   &   0.255 \\
\hspace{-1mm}Inter-set stability        & 26.01           & 0.469        & n/a  & n/a\\
\bottomrule
\end{tabular}
\caption{The results of stability analysis experiment: BLEU, METEOR (MET.), frequency of the class flip (c.f.) and drop of predicted class probability ($\Delta$p.).}
\label{tab:stability}
\end{table}

\paragraph{Masking input image}
We asked 
annotators to \emph{cover} with white rectangles parts of images that contained the decision rationale indicated in the NLE, 50\% area at most (see App.~\ref{app:covering} for details).
We re-classified covered images and measured changes in prediction and the average decrease in the probability of the originally predicted class.
We also performed an opposite experiment, with the annotators \emph{highlighting} only parts of image mentioned in the explanation and covering the rest.

For covering, the use of our NLEs resulted in the highest average probability decrease and the change of the original prediction for 88\% of examples (see Table~\ref{tab:intervention}). 
Our method reached the best results in the highlighting experiment as well, producing the least amount of changes.

We also re-ran our NLE pipeline with parts of the input image covered. This led to significant changes: on average, 78\% (median 90\%) of the neurons indicated in MRs were different.

\paragraph{Masking influential neurons}
To show the NLEs' ability to reflect classifier decisions, we asked the annotators to read the NLEs and select up to five most influential neurons from a MILAN-annotated list. 
The classifier was then re-run with the selected neurons masked. The results 
in Table~\ref{tab:intervention} reveal that masking neurons suggested by our NLEs led to a five times higher decrease in the predicted class probability and over three times higher class flip rate than baselines. 

\begin{figure*}[!ht]
    \includegraphics[width=0.45\textwidth]{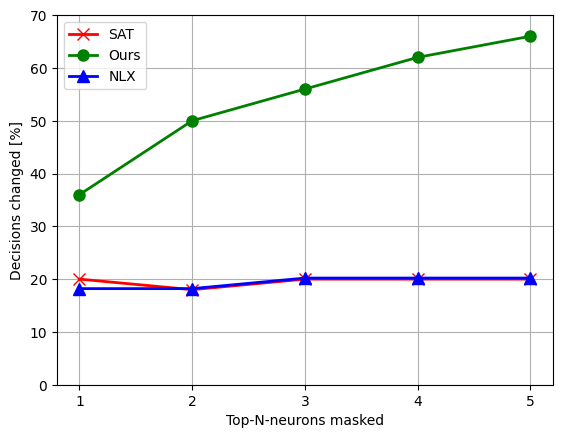}
\hfill
    \includegraphics[width=0.45\textwidth]{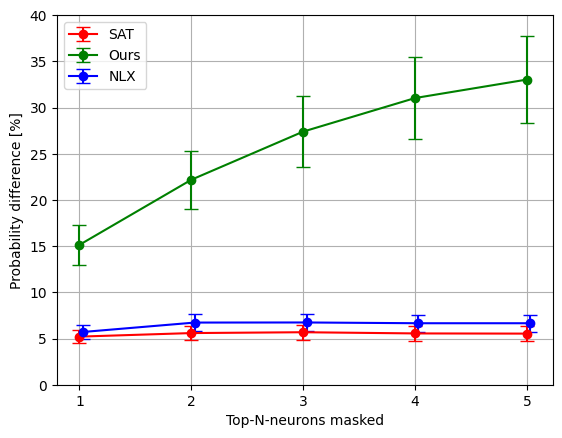}
\hfill
    \caption{Results of neuron masking experiments when varying the number of masked neurons. Left: percentage of changed predictions after masking selected neurons; right: decrease of the probability of the predicted class after masking selected neurons.}
    \label{fig:masking}
\vspace{5mm}
\small
    \begin{tabular}{p{0.22\textwidth}p{0.20\textwidth}p{0.5\textwidth}}

\toprule
\textbf{Image} & \textbf Class & \textbf{Explanation} \\
\midrule 
\raisebox{-0.8\totalheight}{\includegraphics[width=0.21\textwidth]{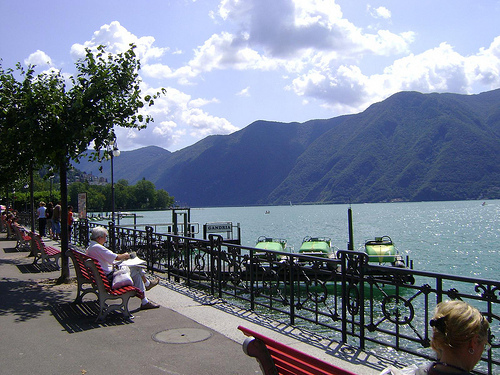}} &Target: "lakeside" \newline Prediction: "lakeside" &The model classified this image as 'lakeside' because it detected water in the lower half and the bottom-right corner of the image. It also noticed volcanoes and mountains in the upper half, and walkways at the bottom-left corner and bottom. Additionally, the model found items with straight features at the left, bottom-left corner, and bottom of the image. \\
\midrule
\raisebox{-.87\totalheight}{\includegraphics[width=0.21\textwidth]{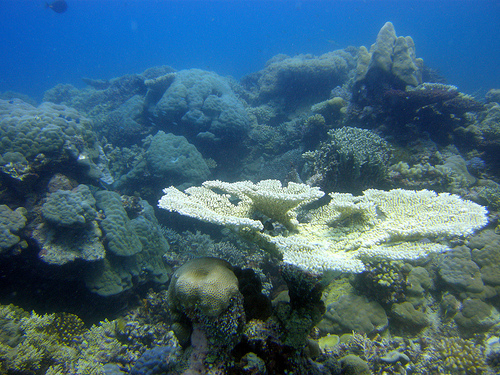}} &Target: "coral reef" \newline Prediction: "coral reef"  & The model classified this image as a "coral reef" due to the detection of aquatic life throughout the image, particularly at the bottom, left, and right. The presence of water covering the entire image and nature-related elements further supported this classification. The model also noticed items that are connected in the lower half of the image and similar color patterns, which are common in coral reef environments. \\
\midrule
\raisebox{-.8\totalheight}{\includegraphics[width=0.21\textwidth]{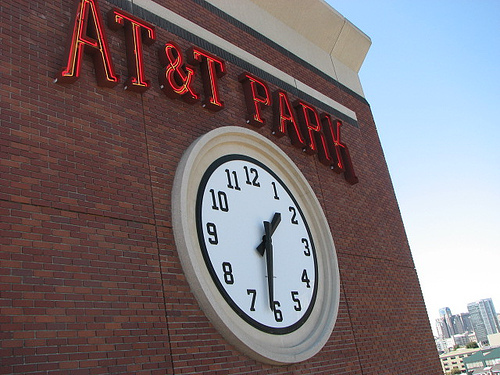}} & Target: "wall clock" \newline Prediction: "wall clock" & The model classified this image as a "wall clock" because it detected clocks and other gauges, circular, round objects, and black and white objects at the bottom-right corner, center, and bottom of the image. It also noticed rounded edges in the right half of the image and the indent in an hourglass shape at the bottom-right corner, center, right, and bottom. \\
\bottomrule
\end{tabular}
\caption{Examples of explanations provided by our method.}
\label{tab:case_study_table}
\end{figure*}

More detailed results are presented in Fig.~\ref{fig:masking}.
Masking neurons in the order indicated by the annotators using our method leads to an increasing change in the classifier's prediction and a gradual decrease in the predicted class probability value.
In contrast, masking the neurons indicated using other methods leads to a small decrease in class probability for one masked neuron and almost no further decrease for more masked neurons. As we attribute the effect of the first masked neuron to examples of classes that are highly related to a singular pattern (e.g., a neuron annotated “water” for the class “sea”), this indicates that annotators gain very little insight into how the neural network made a decision from the explanations provided by the baselines.

\paragraph{Explanation stability analysis}
Following~\citet{wiegreffe-etal-2021-measuring},
we measure explanation robustness against adding random noise to the input image (intra-set stability, see  App.~\ref{app:noise} for details) by comparing BLEU~\cite{papineni-etal-2002-bleu}, METEOR~\cite{lavie-agarwal-2007-meteor} as well as class flip frequency and class probability change against the original predictions.
We also check for outputs' diversity (inter-set stability) using BLEU  and METEOR  overlap  against explanations for other classes.
The results in Table~\ref{tab:stability} show that the explanations are both distinct for different classes and highly sensitive to noise:
As we add more noise resulting in increased classification changes,
the explanation BLEU and METEOR gradually drop, but they are at their lowest when comparing between different classes.

\paragraph{Reliability of the MR-to-text transform}
The human evaluation of our approach's MR-to-text reliability was similar to plausibility evaluation, but limited to non-expert Prolific annotators.
We asked five yes-no questions on information in the text not grounded in the MR (i.e., hallucinations), omission of MR information, fluency, spatial information fidelity,
and overall correctness (see App.~\ref{app:mrtotext}).


The results show the MR-to-text conversion as highly reliable, as only 8\% texts contain hallucinations. Omissions are more frequent (44\%), but this is expected as the LLM is instructed to summarise the MR and prioritise readability. 
This factor most likely affected the overall score (58\%). 
The explanations are mostly fluent (96\%), with correct spatial information (82\%).

Additionally, we repeated this experiment for explanations generated by an open LLM (Llama 3 70B) instead of GPT-4.
The results presented in App.~\ref{app:mrtotext} show that open LLM generated NLEs with a significantly higher number of hallucinations and omissions, but this did not affect the overall quality score given by the annotators.



\section*{Acknowledgements}
Co-funded by the European Union (ERC, NG-NLG, 101039303) and National Science Centre, Poland (Grant No.~2022/47/D/ST6/01770). 
This work used resources of the LINDAT/\hspace{0mm}CLARIAH-CZ Research Infrastructure (Czech Ministry of Education, Youth, and Sports project No.~LM2018101).
For the purpose of Open Access, the author has applied a CC-BY public copyright licence to any Author Accepted Manuscript (AAM) version arising from this submission.

\section*{Limitations}
This paper produces a new method for plausible and more faithful natural language explanations for image classification.
Although we believe that the method provides significantly better  faithfulness than the previously proposed methods, it does not obtain completely faithful explanations. 
The faithfulness of the explanations provided by our method depends on the quality of the neuron annotations produced by MILAN and the neurons indicated by LRP. Both techniques can be considered as state of the art, but they still occasionally produce incorrect results.
Therefore, the results of NLE methods should be treated with caution.
Additionally, this work uses pre-trained language models, which are known to expose certain social biases reflected in their training data.

\bibliography{anthology,custom}

\appendix

\section{Examples of explanations}
\label{app:examples}
Several examples of explanations provided by the methods under study are given in Tab.~\ref{tab:model_explanations}.

\begin{table*}[h]
\centering
\small 
\begin{tabular}{c|p{6.5cm}|p{4cm}|p{3cm}}
\toprule
& {{Our Method}} & {{NLX-GPT}} & {{Show, Attend and Tell}} \\
 & & & \\
\midrule
1 & The model classified this image as a "cradle" due to the detection of items with circular features on the right half of the image, items with straight features in the center and on the entire right side, and rounded edges in pictures at the center, right, and bottom of the image. It also noticed the indent in an hourglass shape in the center, entire right, and bottom of the image. Additionally, human hands were detected at the center, right, and bottom of the image. & There is bathroom in the image because there is a sink and a toilet. & A black and white photo of a guitar case. \\
\midrule
2 & The model classified this image as "volcano" because it detected elements of nature in the lower half of the image. It also identified a gusher of water, which could be interpreted as lava, across various parts of the image including the left, center, right, and bottom. Additionally, the model found items with both curved and straight features in the lower half of the image. & There is mountain in the image because there is a large mountain in the background. & A view of a mountain range in a cloudy sky. \\
\midrule
3 & The model classified this image as a "library" because it detected shelves and books in the lower half of the image. It also noticed objects with led, text, and circular objects, rectangular objects, and cubed objects throughout the entire image. Additionally, items with straight features were found in the left half of the image, and grids were seen in the bottom-right corner and the right side of the image. & There is library in the image because there are bookshelves full of books. & A bookshelf filled with lots of books. \\
\bottomrule
\end{tabular}
\caption{Examples of image classification explanations provided by method under study.}
\label{tab:model_explanations}
\end{table*}


\section{Simplification rules for spacial positions}
\label{app:rules}
The spatial information about neuron activation is learned by inspecting activation maps of the chosen neurons. The activation map is converted to a binary image, such that if a given pixel of an activation map exceeds half of a maximal value in that activation map it is converted to the value of one, and the rest of the pixels are assigned with the value zero. Then the binary activation map is divided into a 3x3 grid of congruent squares oriented such that the ordering begins with a 0 in the top-left corner, progressing sequentially across each row from left to right and top to bottom, culminating with the number 8 in the bottom-right position.  The basic positions names are given below.
\lstset{captionpos=b}
\par
\begin{minipage}{\linewidth}
\begin{lstlisting}[label=lst:basic_positions]
0 : "top-left corner", 
1 : "top", 
2 : "top-right corner", 
3 : "left", 
4 : "center", 
5 : "right", 
6 : "bottom-left corner", 
7 : "bottom", 
8 : "bottom-right corner"
\end{lstlisting}
\end{minipage}
The final appearance of the positions appended to the meaning representation is governed by a set of rules, where position names are assigned to sets of basic positions.
 If all basic positions from the set are present in the MR, we replace them with a corresponding compound position.  These sets and position names are given below.
\par
\begin{minipage}{\linewidth}
\begin{lstlisting}[label=lst:compound_positions,breaklines=true]
"entire top": {0, 1, 2},
"entire bottom": {6, 7, 8},
"entire left": {0, 3, 6},
"entire right": {2, 5, 8},
"perimeter": {0, 1, 2, 5, 8, 7, 6, 3},
"center cross": {1, 3, 4, 5, 7},
"upper half": {0, 1, 2, 3, 4, 5},
"lower half": {3, 4, 5, 6, 7, 8},
"left half": {0, 1, 3, 4, 6, 7},
"right half": {1, 2, 4, 5, 7, 8}
\end{lstlisting}
\end{minipage}
Additionally if "entire *" and "* half" coexist in the meaning representation simultaneously, where "*" represents one of \{"left", "right", "top", "bottom"\}, in both compound positions names, the "entire *" is deleted from the meaning representation, since it is comprised of the subset of the set, which makes up corresponding "* half". An example would be if "entire top" and "upper half" are both in meaning representation, the "entire top" would be deleted since it is comprised of \{0, 1, 2\} tiles, which also partially make up the "upper half" (\{\textbf{0}, \textbf{1}, \textbf{2}, 3, 4, 5\}). Finally, if more than any, distinct, 6 out of 9 tiles are active, every position of a given neuron is deleted and the spatial information placeholder is set to "entire image". This whole approach significantly reduces the length of the positional representation, while simultaneously making the position names more plausible to the system user, since we believe the final space of possible positions is natural and intuitive for humans to grasp quickly and effectively.

\section{Prompt for MR-to-text conversion}
\label{app:prompt}
To convert meaning representations into text, the following prompt was applied to the language model:

{\small
\begin{spverbatim}
You are given a problem of creating textual explanation of an image classification performed by neural network. You will be given a Python object representing network output in the form of `'image class', [('detected object', 'position'), ...]`. I want you to convert this object into a textual explanation. You should:
1. Create a grammatically correct sentence which will explain the model's decision.
2. Decide which detected objects do not fit with image class and do not include them in the explanation. For instance, 'dentist' class and 'animal heads' objects are completely unrelated. However, the descriptions that aren't directly related to image class, but can be indirectly correlated, especially in terms of shape, color, or texture resemblance should be included (like 'fountain' and 'sea' because of the water they have in common or 'brick wall' and 'grid' because the texture is similar). Never mention that you chose neuron descriptions and do not talk about the neuron descriptions that were discarded.
3. Prioritize the readability of the explanation. Include only essential detected objects and aggregate information to shorten the explanation.
4. Aggregate positions if possible, for example ['bottom-left corner', 'bottom', 'bottom-right corner'] should be aggregated into 'bottom'. If the positions list is too long or too ambiguous do not include them in the explanation. 

Here is an example.
Python object:
“lakeside, [{'description': 'Nature', 'positions': ['left', 'right', 'bottom']}, {'description': 'The sky', 'positions': ['top-right corner', 'bottom-left corner', 'bottom']}, {'description': 'Red and white colored objects', 'positions': ['left', 'right', 'bottom']}, {'description': 'The ocean', 'positions': ['left', 'right', 'bottom']}, {'description': 'Animal heads', 'positions': [], 'id'}, {'description': 'The color red', 'positions': ['left', 'right', 'bottom']'}, {'description': 'White backgrounds', 'positions': ['bottom', 'left']}, {'description': 'Grass', 'positions': ['left', 'bottom-left corner', 'bottom', 'bottom-right corner']}, {'description': 'Dogs and guinea pig', 'positions': ['center']}, {'description': 'The color green', 'positions': ['top-right corner', 'right', 'bottom-right corner']]”
Answer: “The model assigned this image to the "lakeside" class because in the last layer it discovered nature, and the ocean at the left, right, and bottom of the image. It also detected grass at the left, bottom-left corner, bottom, and bottom-right corner and the color green at the right of the image.
(...)
\end{spverbatim}
}

\section{Human evaluation of explanation plausibility}
\label{app:plaus}
The annotators are presented with an image, model's prediction and a natural language explanation. 
Each question is answered on a scale from 1 (low) to 5 (high).
The following questions are asked:
\begin{itemize}
    \item How fluent (linguistically correct) the text is?	
    \item How easy to understand the text is?	
    \item How convincing do you find the explanation of the decision made by the model?	
    \item After reading the explanation, how well do you understand how the decision of the model was taken?	
    \item How would you rate the overall quality of the explanation?

\end{itemize}
The annotation instructions are provided in the code repository.

\section{Human evaluation of MR-to-text transformation}
\label{app:mrtotext}
The annotators are presented with a meaning representation in the form of formatted JSON without a given image, since it should not influence the assessment of MR-to-text transformation. The following binary questions are asked:
\begin{itemize}
    \item Does the text contain information that was not present in the meaning representation?	
    \item Is there any important information from meaning representation omitted in the text? 
    \item Is the text linguistically correct?	
    \item If any spatial compression occurred between explained neurons, is the said compression correct?
    \item 
    Overall, do you find this meaning representation to text transformation acceptable, i.e. sufficiently good for explanation purposes?

\end{itemize}
The annotation instructions are provided in the code repository.

The experiment was carried out on explanations generated with two LLMs: GPT-4 (used in all other experiments) and an open-weight alternative, namely Llama 3 70B from \texttt{ollama} library\footnote{\url{https://ollama.com/library/llama3}}. Both sets of explanations were generated using the same prompt and for the same MRs.

The results are presented in Tab.~\ref{tab:mrtotext}. Both GPT-4 and Llama appear to have a similar ability to perform spatial compressions, but for the other factors examined, the NLEs generated by Llama fall short of those generated by GPT-4.
Llama's explanations are almost 6 times more likely to contain hallucinations. They also have more omissions and lower fluency. Nevertheless, the overall correctness of the NLE's generated by both LLMs is the same and not overly high. 
We think that this result is influenced by the fact that some annotators penalise the correctness of NLE's too much due to omitted information from the meaning representation. Note that the MRs are quite long and the generated NLEs are supposed to summarise them and shorten them to improve readability, thus omitting some information.

\begin{table}
\small\centering
\begin{tabular}{lrr}
\toprule
Question &  GPT-4 & Llama 3 70B\\
\midrule
Hallucinations $\downarrow$     & 0.08 & 0.46 \\
Omissions      $\downarrow$     & 0.44 & 0.64 \\
Fluency       $\uparrow$  & 0.96  & 0.86\\
Spacial compression $\uparrow$ & 0.82 & 0.82 \\
Overall correctness $\uparrow$ & 0.58 & 0.58\\
\bottomrule
\end{tabular}
\caption{Human evaluation of MR to text conversion with GPT-4 and Llama 3 70B as backbone LLMs. The percentage of "yes" answers is reported. }
\label{tab:mrtotext}
\end{table}

\section{Details on stability experiments}
\label{app:noise}
Let us assume that an image is a matrix $X$, such that $ X \in [0, 1]$. We model input perturbations by adding random noise to the images, sampled from the standard normal distribution. 
Since an interval of possible pixel values  is $[0, 1]$, to account for the unboundedness of a standard normal distribution we use a clipping operation defined as follows:
\begin{equation*}
\text{clip}(x) = 
\begin{cases}
0, & \text{if } x < 0 \\
x, & \text{if } 0 \leq x \leq 1 \\
1, & \text{if } x > 1
\end{cases}
\end{equation*}
The formula for the function $\zeta$ that adds a perturbation to the image is given below.
\begin{equation*}
\zeta(X)=clip(\mathcal{N}(0, 1) \cdot i + X)
\end{equation*}
where $i$ is the noise intensity.
\par
We perform two types of stability experiments:
\begin{enumerate}
    \item \textit{intra-set stability}: We compare pairs of unaltered images ($1^{st}$ set of images) and corresponding images mapped by the $\zeta$ function ($2^{nd}$ set of images). After the $\zeta$ mapping is performed, we run the proposed method on both images, yielding two explanations. We then compute various language similarity metrics between the two explanations. 
    
    We compare two kinds of pairs of images: unaltered - lightly perturbed ($i=0.05$)  and unaltered - heavily perturbed ($i=0.2$). 
    \item \textit{inter-set stability}: We compare explanations pairs produced for different images to verify the diversity of generated explanations. 
\end{enumerate}
The computations were conducted on explanations produced by the proposed method for a 500-element subset of validation ImageNet data.
The subset was constructed by randomly picking 50 examples from 10 selected, diverse classes (library, over skirt, palace, prison, wall clock, lakeside, coral reef, volcano, fountain, basset) in a stratified manner.

\begin{figure}
    \centering\small 
    \includegraphics[width=.4\textwidth]{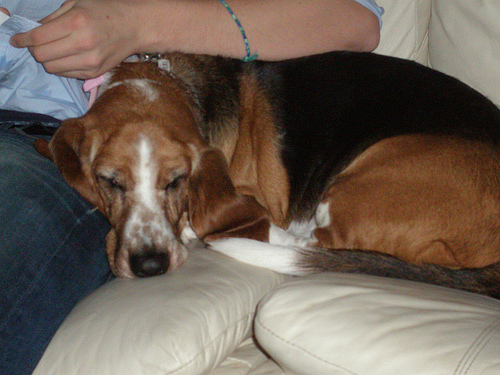}
    Prediction: "basset"\\
    \includegraphics[width=.4\textwidth]{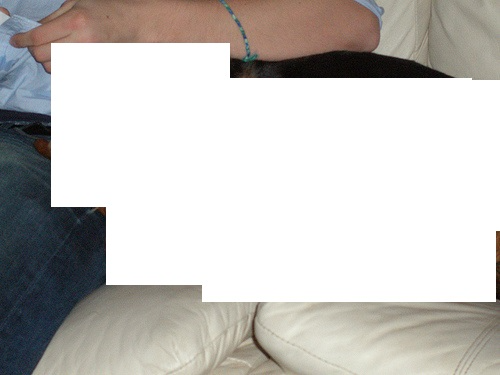}
    Prediction: "studio couch"\\
    \caption{Example image with and without human annotation in covering experiment.}
    \label{fig:covering}
\end{figure}

\section{Details on the covering experiment}
\label{app:covering}
Given an image and explanation, we instructed the annotator to cover the decision rationales with white rectangles.
The annotator instruction is given below:
\begin{quote}
    Based on the following explanation of the classifier's prediction, cover the reasons for its decision with white rectangles. You can use as many rectangles as you like, but the total area of the covered image cannot be larger than half of the image. If it is not possible to cover the mentioned reasons by covering only 50\% of the image, please do your best to cover the most important information. 
\end{quote}
An example of annotation provided is given in Fig~\ref{fig:covering}.

\section{Layer-wise Relevance Propagation }
\label{app:lrp}
To choose the most relevant neurons, the well-established Layer-wise Relevance Propagation (LRP) method is applied~\cite{Bach2015OnPE}. LRP performs a special backward pass through the neural network  
to establish the influence of each neuron to the final prediction.
Starting with the predicted value, LRP distributes it among the neurons in each layer, assigning them relevance scores.
The following rule for relevance reallocation is used: 
$$    R_i = \sum_j \frac{z_{ij}}{\sum_i {z_{ij}}} R_j
$$
where $R_i$ is the $i$-th neuron relevance score, 
$z_{ij}$ express how much $i$-th neuron has contributed to make $j$-th neuron relevant (calculated as the product of the neuron's activation and the corresponding weight), the sums $\sum_i$ ($\sum_j$) iterate over all neurons in a given (next) layer.

\paragraph{Choice of LRP as an explanation method}
Although we present a pipeline approach and there is some variability in how it can be implemented, we believe there are important reasons for using our pipeline with LRP.

First, unlike many other methods, LRP works at the neuron level, which is strictly required by our method. Therefore, methods that provide pixel-level importance scores such as RISE~\cite{petsiuk2018rise} or Grad-CAM~\cite{selvaraju2017grad}, methods that typically work on image segments such as LIME~\cite{lime} or SHAP~\cite{NIPS2017_7062} are not suitable for our approach.
 
Second, LRP has been shown to achieve high faithfulness in many studies and can be considered state of the art in this respect. Note that achieving high faithfulness is the main goal of our approach.

Finally, LRP is theoretically motivated, has been shown to be useful in many applications, and has stable open source implementations. 

\section{Statistical analysis of the human evaluation results}
\label{app:stat}
For the results of human evaluation of plausibility, we performed the non-parametric global Friedman test followed by Nemenyi post-hoc analysis (as recommended in \cite{JMLR:v7:demsar06a}). 
We were able to reject the null hypothesis of the Friedman test for all the measures with p<0.001. The Nemenyi post-hoc analysis with $\alpha=5\%$ confirmed that our method obtains statistically significant improvements over other compared methods on both plausibility measures and the overall quality measure. On the fluency measure, our method is undistinguishable from SAT.

\begin{figure}
    \centering
    \begin{subfigure}{0.4\textwidth}
    \includegraphics[width=\textwidth]{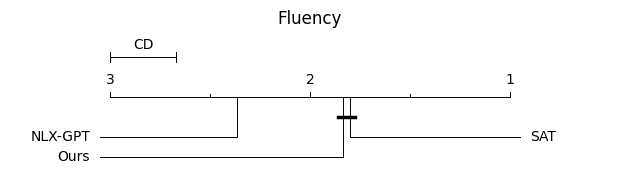}
    \caption{Fluency}
    \end{subfigure}
\hfill
\begin{subfigure}{0.4\textwidth}
    \includegraphics[width=\textwidth]{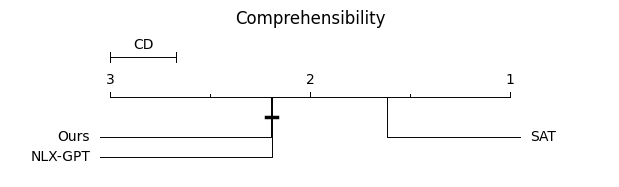}
    \caption{Comprehensibility}
    \end{subfigure}
\hfill
\begin{subfigure}{0.4\textwidth}
    \includegraphics[width=\textwidth]{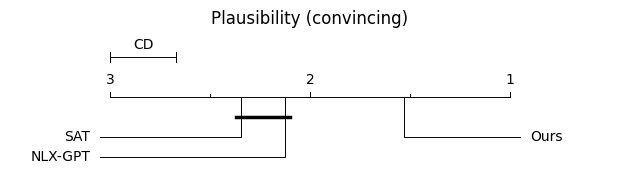}
    \caption{Plausibility (convincing)}
    \end{subfigure}
\hfill
\begin{subfigure}{0.4\textwidth}
    \includegraphics[width=\textwidth]{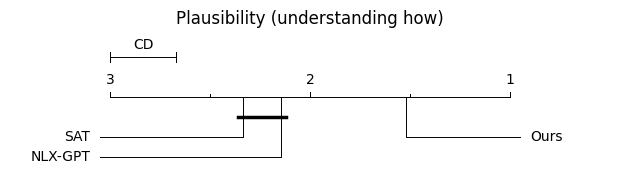}
    \caption{Plausibility (explanatory)}
    \end{subfigure}
\hfill
\begin{subfigure}{0.4\textwidth}
    \includegraphics[width=\textwidth]{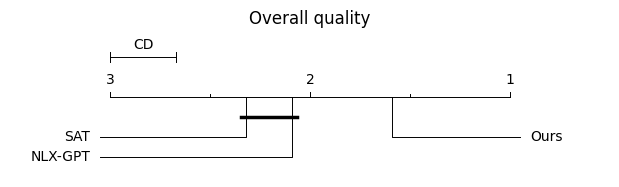}
    \caption{Overall quality}
    \end{subfigure}
\hfill

    \caption{The results of Nemenyi post-hoc analysis for different aspects of evaluated explanations.}
    \label{fig:nemenyi}
\end{figure}

The critical distance plots from Nemenyi post-hoc analysis are provided in Figure~\ref{fig:nemenyi}. The lower result, the better. If the difference between the methods is not statistically significant, their results are connected with a thick horizontal line. More details on these plots can be found in~\cite{JMLR:v7:demsar06a}.

\end{document}